\documentclass[sigconf]{acmart}
\usepackage[]{algorithm2e} 
\usepackage{algorithmic} 
\usepackage{amsmath}
\usepackage{amsthm}
\usepackage{bbm}
\usepackage{multirow}

\DeclareMathOperator*{\argmin}{arg\,min}

\newcommand{\cmmnt}[1]{\ignorespaces}
\AtBeginDocument{%
  \providecommand\BibTeX{{%
    \normalfont B\kern-0.5em{\scshape i\kern-0.25em b}\kern-0.8em\TeX}}}

\acmConference[]{}{}{}
\begin{document}

\title{Joint Analysis of Single-Cell Data across Cohorts with Missing Modalities}

\author{Marianne Arriola}
\affiliation{%
  \institution{Department of Computer Science, Cornell Tech, Cornell University}
  \city{New York}
  \state{NY}
  \country{USA}
  \postcode{10065}
}
\email{ma2238@cornell.edu}

\author{Weishen Pan}
\affiliation{%
  \institution{Department of Population Health Sciences, Weill Cornell Medicine, Cornell University}
  \city{New York}
  \state{NY}
  \country{USA}
  \postcode{10065}
}
\email{wep4001@med.cornell.edu}

\author{Manqi Zhou}
\affiliation{%
  \institution{Department of Computational Biology, Cornell University}
  \city{Ithaca}
  \state{NY}
  \country{USA}
  \postcode{14853}
}
\email{mz335@cornell.edu}

\author{Qiannan Zhang}
\affiliation{%
  \institution{Department of Population Health Sciences, Weill Cornell Medicine, Cornell University}
  \city{New York}
  \state{NY}
  \country{USA}
  \postcode{10065}
}
\email{qiz4005@med.cornell.edu}

\author{Chang Su}
\affiliation{%
  \institution{Department of Population Health Sciences, Weill Cornell Medicine, Cornell University}
  \city{New York}
  \state{NY}
  \country{USA}
  \postcode{10065}
}
\email{chs4001@med.cornell.edu}

\author{Fei Wang}
\affiliation{%
  \institution{Department of Population Health Sciences, Weill Cornell Medicine, Cornell University}
  \city{New York}
  \state{NY}
  \country{USA}
  \postcode{10065}
}
\email{few2001@med.cornell.edu}

\renewcommand{\shortauthors}{Trovato and Tobin, et al.}
\newcommand{\methodname}{$SC^5$}
\newcommand{\methodnamerm}{$\rm SC^5$}

\begin{abstract}
Joint analysis of multi-omic single-cell data across cohorts has significantly enhanced the comprehensive analysis of cellular processes. However, most of the existing approaches for this purpose require access to samples with complete modality availability, which is impractical in many real-world scenarios. In this paper, we propose \methodname~ (Single-Cell Cross-Cohort Cross-Category) integration, a novel framework that learns unified cell representations under domain shift without requiring full-modality reference samples. Our generative approach learns rich cross-modal and cross-domain relationships that enable imputation of these  missing modalities. Through experiments on real-world multi-omic datasets, we demonstrate that \methodname~ offers a robust solution to single-cell tasks such as cell type clustering, cell type classification, and feature imputation. 
\end{abstract}




\maketitle

\section{Introduction}

Advancements in multi-omics single-cell high-throughput sequencing technologies have introduced novel approaches to integrate rich information across modalities. Techniques such as single-cell RNA sequencing (scRNA-seq) combined with the assay for transposase-accessible chromatin using sequencing (ATAC-seq) allow for the concurrent analysis of the transcriptome and chromatin accessibility within the same cell\cite{buenrostro2015single}. Additionally, Cellular Indexing of Transcriptomes and Epitopes by Sequencing (CITE-seq) enables the measurement of surface protein and transcriptome data within individual cells using oligonucleotide-tagged antibodies\cite{stoeckius2017simultaneous}. By integrating data from diverse omics, including the transcriptome, proteome, and epigenome, we can achieve a more comprehensive understanding of genome regulation from various perspectives\cite{xu2022smile,gayoso2021joint,zhou2023moetm, ashuach2023multivi,cao2022pamona,demetci2022scotv2,baysoy2023technological}.

\begin{figure}
    \begin{center}
        \includegraphics[scale=0.45]{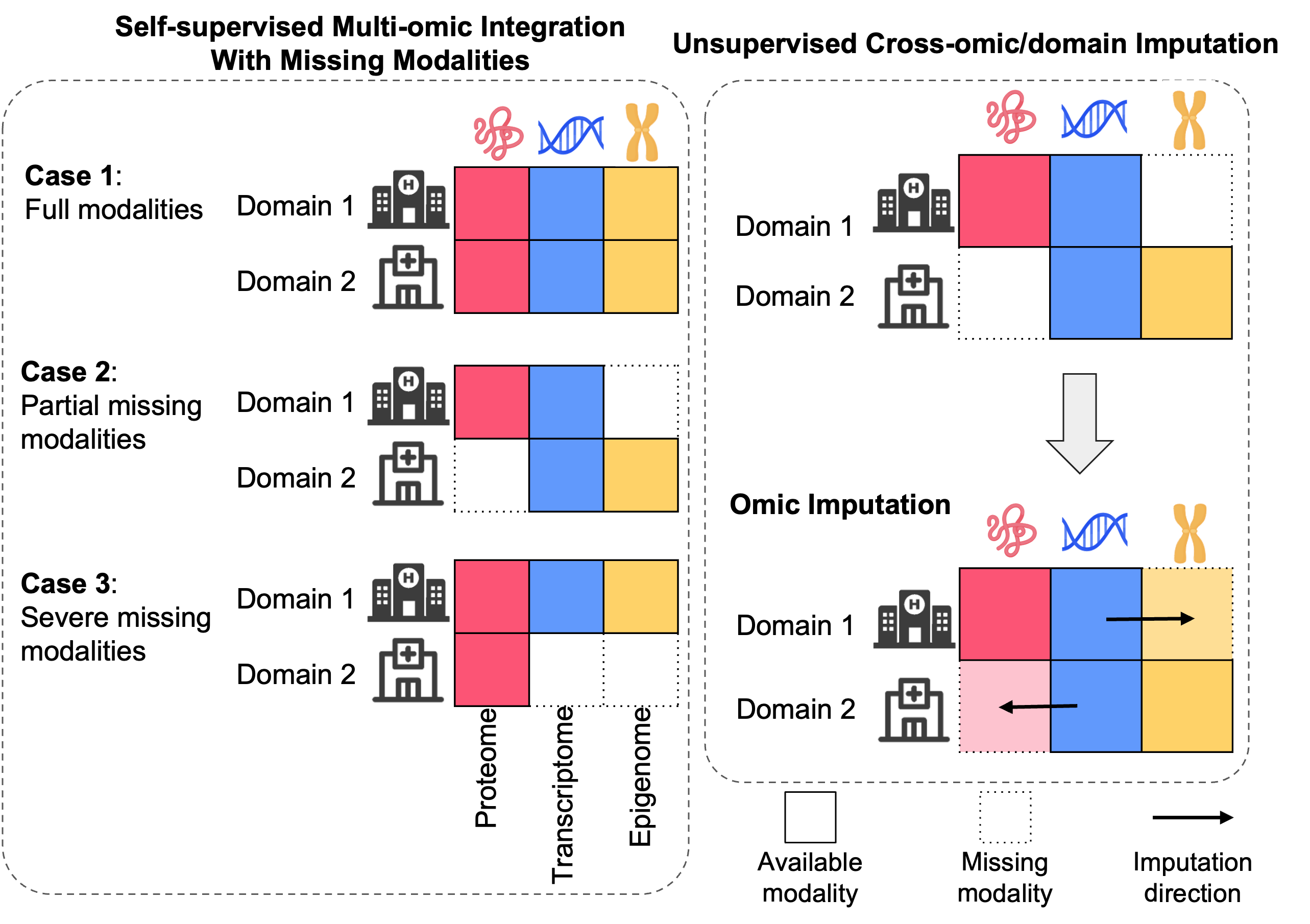}
    \end{center}
    \caption{Problem overview. \textit{Left:} Different cases of modality availability. \textit{Right:} Different cases of modality imputation.}
    \label{fig:missing_modality_task}
    \vspace{-1.0em}
\end{figure}

Despite the great efforts on single-cell integration, integrating multi-omic data across cohorts from different datasets remains a relatively underexplored problem. To make it consistent and more generalizable, we treat each cohort from each dataset as a specific domain and each omic as a specific modality. This problem becomes even more challenging due to the incompleteness of omics in some cohorts, as shown in Figure \ref{fig:missing_modality_task}. This setting is also referred to as cross-cohort cross-category ($C^4$) learning \cite{rajendran2024learning}. Most current single-cell integration methods either lack mechanisms to learn from \textit{incomplete multi-omic datasets} or they assume the existence of reference samples with complete modalities\cite{zhou2023moetm, braams:babel,gong2021cobolt}. When some modalities are totally missing in some/all datasets, the lack of mechanisms to address this gap overlooks the opportunity to leverage the rich cross-modality interactions present in paired samples. Similarly, when real-world datasets lack reference samples with full modality coverage\cite{rajendran2023patchwork, Lotfollahi2022multigrate} (Case 1 in Figure 1), the presumption of a complete reference dataset becomes untenable. Moreover, the discrepancy of the feature distributions across datasets, which is also called batch effect in single-cell settings, is a critical issue for single-cell integration to be addressed \cite{luecken2022benchmarking}. In this work, we consider the setting of integrating multi-omics data under batch effect across cohorts where modalities may be entirely missing within a domain, departing from previous notions of modality scarcity.

We propose \methodname~(Single-Cell Cross-Cohort Cross-Category) to jointly analyze single-cell data across different domains under missing modality settings (Cases 2 and 3 in Figure \ref{fig:missing_modality_task}). At a high level, our framework models latent topics underlying single-cell multi-modal features in each domain. We implement a variational autoencoder that learns modality-dependent and domain-dependent topic feature distributions in an unsupervised manner. In particular, we model the data as being generated by a combination of domain-dependent and modality-dependent latent factors. These domain-dependent factors describe shared information across modalities within each domain. On the other hand, the modality-dependent factors describe shared information across domains with common modalities. This information sharing across modalities and domains enables feature learning that is robust in the setting where full-modality samples are unavailable. \methodname~ can transfer rich multimodal features between domains and impute missing modalities using observed modalities. The contributions of this proposed framework are:

\begin{enumerate}
    \item Modality-invariant feature learning without requiring full modality availability.
    \item Adaptation of disparate domain-specific, multi-modal feature distributions under missing modality settings.
    \item Imputation of modality features that are entirely unseen in a domain.
\end{enumerate}

Through experiments on real-world datasets, we show that \methodname~ can effectively learn cross-modality relationships and thus infer more representative single-cell embeddings even under missing modality settings. With both qualitative analysis and quantitive metrics, we demonstrate that our approach outperforms the baselines in different tasks. Code is available at \url{https://github.com/anonsc5kdd/sc5}.

\section{Related Works}

\subsection{Integration with Missing Modalities}

Recent methods in multi-modal integration under missing modality settings perform imputation as a pre-processing step.
SMIL \cite{ma2021smil} addresses bias from modality scarcity by performing missing modality imputation before learning joint embeddings. MultiModN \cite{swamy2023multimodn} proposes a multimodal fusion architecture that maintains robustness to missing modalities by sequentially encoding each modality, skipping over missing samples. Ding {\em et. al} \cite{ding2015missinglowrank} use knowledge transfer to produce embeddings for a missing modality in a target domain by using an available common modality to learn low-rank factors. Konwer {\em et. al} \cite{konwer2023mamlbrain} propose an adversarial architecture that learns modality-agnostic embeddings by using a discriminator to predict the absence/presence of modalities. However, their approach requires their discriminator to be trained over $2^M-1$ possible missing modality combinations for $M$ modalities. 

Another line of recent methods apply multi-modal integration for single-cell data under missing modality settings but require full-modality reference samples. Multigrate \cite{Lotfollahi2022multigrate} uses a product-of-experts encoder to integrate available modalities and a condition vector to incorporate batch information. However, the conditional autoencoder has limited interpretability due to the neural network decoder. Cobolt \cite{gong2021cobolt} is a hierarchical Bayesian generative model that handles missing modalities by estimating posteriors for each unique modality availability setting. scMoMAT \cite{zhang2023scmomat} pre-computes missing features before using matrix factorization in the integration step. moETM \cite{zhou2023moetm} deals with missing modalities by training a model to reconstruct one modality from another. Most of these methods require samples with full modality availability at training. UINMF \cite{kriebel2022uinmf} can integrate features from single-cell datasets containing shared and unshared features using non-negative matrix factorization. However, it is not designed to jointly leverage multiple modalities from the same cell, integrate across cohorts, or perform missing modality imputation.

\subsection{Integration under Batch Effect}

There are some existing works on addressing batch effect for single-cell integration \cite{haghverdi2018batch, li2020deep}. Li {\em et al.} propose an unsupervised deep embedding algorithm to remove batch effect by clustering \cite{li2020deep}. Haghverdi {\em et al.} propose to use matching mutual nearest neighbors for batch effect removal. However, these techniques to address batch effect are mostly specific to the structure of integration model. Furthermore, they assume access to full-modality data.

\section{Problem Settings}
\subsection{Notations}
We consider the single-cell data from $D$ different domains (i.e. cohorts, hospitals) and $M$ modalities which might be measured in at least one of these domains. For domain $d$, there are $N^{(d)}$ cells measured on available modalities $\mathbf{m}^{(d)}$ ($\mathbf{m}^{(d)} \subseteq \{1,..., M\}$ ). We denote $M_d := |\mathbf{m}^{(d)}|$ as the number of available modalities in domain $d$. We assume all samples in each domain $d$ have corresponding measurements for each available modality in $\mathbf{m}^{(d)}$. For a modality which is not in $\mathbf{m}^{(d)}$, it is completely missing in domain $d$ (there exist no partial samples). The activity of the cell is independently measured by each modality $m$ ($m \in \mathbf{m}^{(d)}$) by a $V^{(m)}$-dim feature vector $\mathbf{X}^{(d, m)}$, where each feature corresponds to unique modality-specific measurements. $\mathbf{x}^{(d, m)}_{n}$ is the value of $\mathbf{X}^{(d, m)}$ for the $n$-th sample in domain $d$. To simplify our notation, we will write $\{\mathbf{x}^{(d,m)}_{n}\}_{m \in \mathbf{m}_d}$ as $\mathbf{x}^{(d, \cdot)}_{n}$ without causing any confusion.



\begin{table}[h]
    \centering
    \caption{Variable Definitions}
    \label{tab:variables}
    \renewcommand{\arraystretch}{1.5}
    \begin{tabular}{|c|p{5cm}|}
    \hline
    \textbf{Variable} & \textbf{Definition} \\
    \hline
    $D \in \mathbb{Z}^+$ & \# domains (e.g. donors/cohorts) \\
    \hline
    $M \in \mathbb{Z}^+$ & \#  modalities in all domains \\
    \hline
    $\mathbf{m}^{(d)}$ & The set of modalities in domain $d$\\
    \hline
    $M^{(d)}$ & Number of modalities in domain $d$ \\
    \hline
    $N^{(d)} \in \mathbb{Z}^+$ & \# cells in domain $d$\\
    \hline
    $V^{(m)} \in \mathbb{Z}^+$& \# modality-specific features \\
    \hline
    $L \in \mathbb{Z}^+$ & Hidden feature dimension \\
    \hline
    $K \in \mathbb{Z}^+$ & \# cross-domain, cross-modality topics \\
    \hline
    \hline
    $\mathbf{X}^{(d, m)}$ & Single-cell features \\
    \hline
    $\boldsymbol{\delta}^{(d, \cdot)}$ & Cell topic embedding \\
    \hline
    $ \theta^{(d)}$ & Cell-topic proportion \\
    \hline
    $\boldsymbol{\alpha} \in \mathbb{R}^{K \times L}$ & Global topic embedding \\
    \hline
    $\boldsymbol{\beta}^{(d, \cdot)} \in \mathbb{R}^{K \times L}$ & Domain-specific topic variation \\
    \hline
    $\boldsymbol{\omega}^{(d, \cdot)} \in \mathbb{R}^{K \times L}$ & Domain-specific topic embedding \\
    \hline
    $\boldsymbol{\rho}^{(\cdot, m)} \in \mathbb{R}^{L \times V^{(m)}}$ & Cross-domain \& modality-specific feature embedding \\
    \hline
    $\boldsymbol{\lambda}^{(d, m)} \in \mathbb{R}^{1 \times V^{(m)}}$ & Sample-specific noise parameter \\
    \hline

    \end{tabular}
    \vspace{-1.0em}
\end{table}

\subsection{Task}

\subsubsection{Learning Integrated Representations}
For a given cell $\mathbf{x}^{(d, \cdot)}_{n}$, we aim to integrate features across the available modalities by learning modality-invariant features $\mathbf{z}^{(d, \cdot)}_{n}$ describing its common cell states.

\subsubsection{Missing modality imputation}
After obtaining the latent variable $\mathbf{Z}^{(d, \cdot)}$ from the observed modalities $\{\mathbf{X}^{(d,m)}\}_{m \in \mathbf{m}^{(d)}}$, we further aim to impute features $\hat{\mathbf{X}}^{(i,\bar{m})}$ for a missing modality $\bar{m}$ in domain $d \in [1, \cdots, D]$.

\section{Method}

We propose \methodname, a novel framework to learn integrated cell embeddings under data heterogeneity by performing information sharing across modalities and domains. We represent cellular processes as being generated by `topics' as in \cite{zhou2023moetm} and implement variational inference to learn the generative parameters and integrated features. To model the discrepancy across different domains, we enable the framework to learn domain-specific topics. Moreover, our design supports imputation of modalities that are entirely missing from a domain. In this section, we first formulate the data generative process from a topic modeling perspective in Section \ref{sec:data-generative-process}. Next, we describe the generative model for integration using a variational approach in Section \ref{sec:model-inference} and how to train this model in Section \ref{sec:train-pro}. In Section \ref{sec:model-ncl}, we incorporate cell-cell similarities as an auxiliary signal during multi-modal integration to preserve modality-specific feature relationships in the embedding space. Finally, we describe how \methodname~ computes integrated representations and performs imputation for missing modalities on testing samples in Section \ref{sec:test-time-inference}.



\subsection{Data Generative Process}
\label{sec:data-generative-process}
\begin{figure}
    \begin{center}
        \includegraphics[scale=0.27]{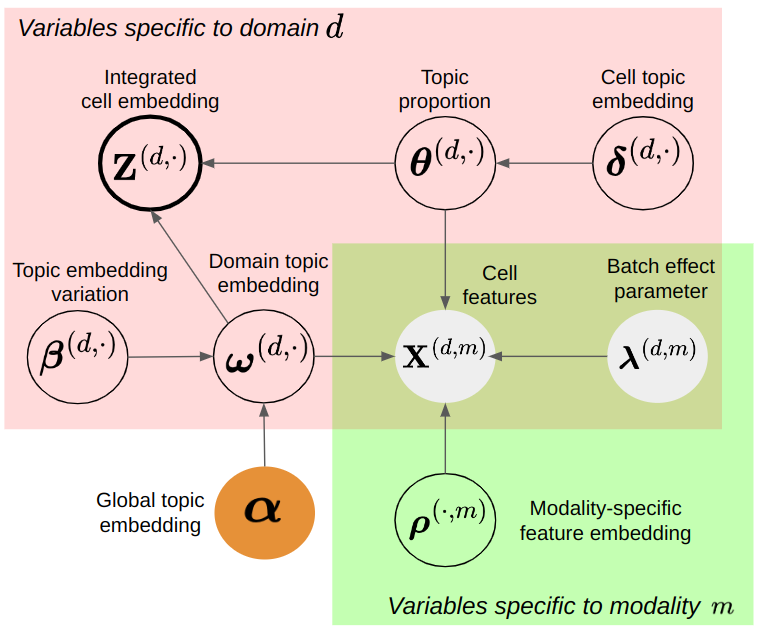}
    \end{center}
    \caption{Graphical illustration of generative model.}
    \label{fig:variables}\
    \vspace{-1.5em}
\end{figure}

\begin{figure*}[tbp]
    \begin{center}
        \includegraphics[scale=0.24]{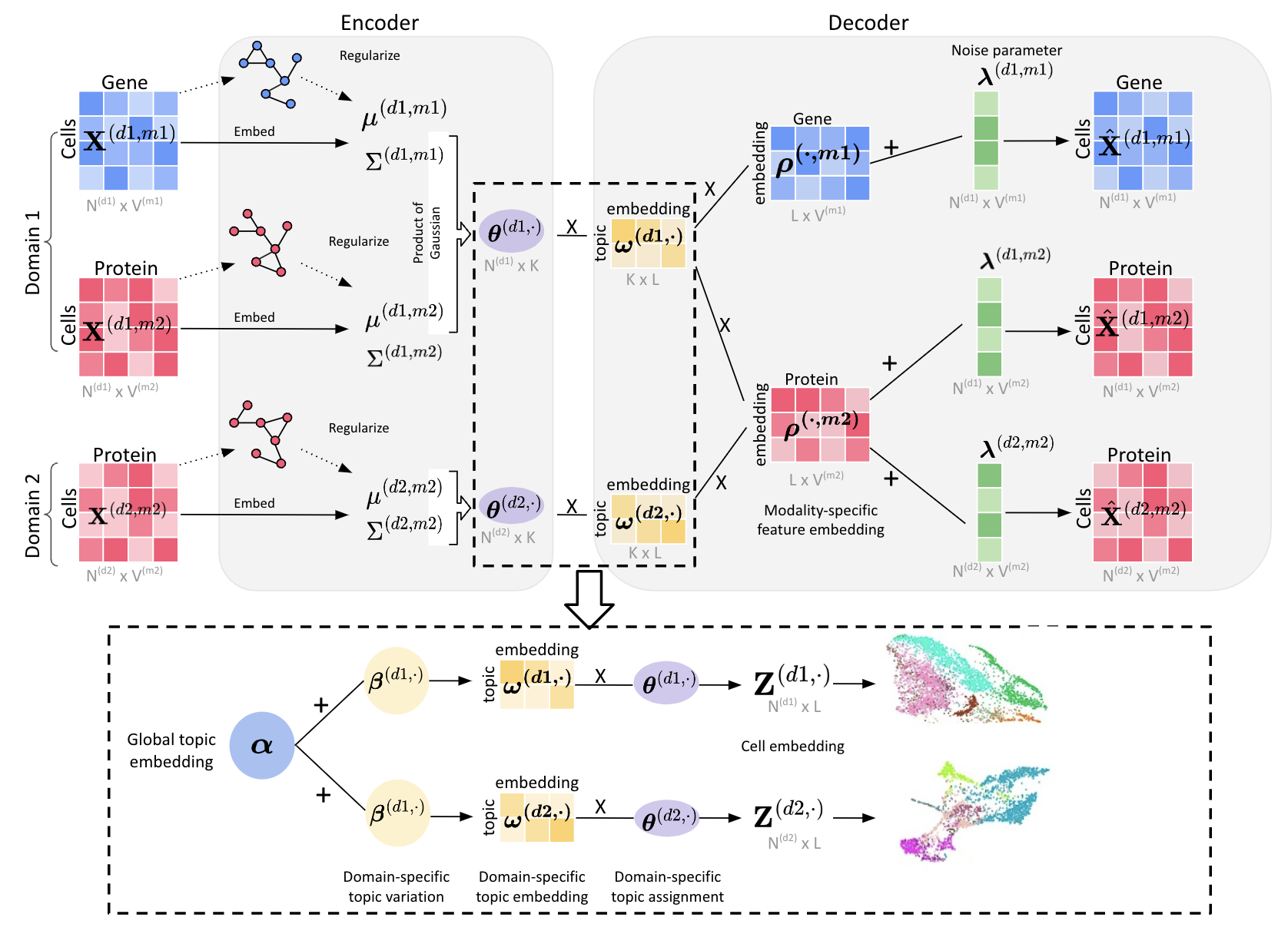}
    \end{center}
    \caption{Model overview. \textit{Top:} The encoder integrates the available modalities for each domain via the product-of-experts (PoE). The decoder reconstructs the modality-specific features by topic and feature embeddings which capture global, domain-dependent, and modality-dependent variation. \textit{Bottom}: Integrated feature representations are obtained by using global and domain-dependent embeddings.}
    \label{fig:model}
    \vspace{-1.0em}
\end{figure*}

The molecular activities in each cell $n$ can be measured with $M$ omics (such as gene expression, protein expression or chromatin accessibility) in a domain $d \in D$. Profiling these omics leads to $M^{(d)}$ feature vectors $\{ \mathbf{x}^{(d,m)}_{n} \}_{m \in \mathbf{m}^{(d)}}$, each with modality-specific feature dimension $V^{(m)}$. We extend \textit{topic modeling}, similar to \cite{zhou2023moetm}, to perform information-sharing between both modalities \textit{and} domains in multi-omic feature integration. We consider each cell as a "document" written in $M$ "languages" (modalities) and by $D$ "authors" (domains). Each feature from the modality $m \in \mathbf{m}^{(d)}$ is a "word" from the $m$-th vocabulary whose vocabulary size is $V^{(m)}$. 
Each sequencing read is a "token" in the document and the abundance of reads mapped to the same feature  is the "word count" in the "document". That is, the value of the $v$-th feature of $\mathbf{x}^{(d,m)}_{n}$ (denoted as $\mathbf{x}^{(d,m)}_{n, v}$) is count of the word $v^{(m)}$ in the "document".

This generative process is described by Figure \ref{fig:variables}. We use $K$ latent topics to describe cells across modalities and domains. We assume these $K$ topics are shared across domains while their proportion distributions might be different across domains. For each cell (indexed by $n$) in  domain $d$, its topic proportion $\boldsymbol{\theta}^{(d, \cdot)}_{n}$ is sampled from a logistic normal distribution:

\begin{equation}
    \boldsymbol{\delta}^{(d, \cdot)}_n \sim \mathcal{N} (\mathbf{M}^{{(d, \cdot)}}, \mathbf{C}^{(d, \cdot)}), \boldsymbol{\theta}^{(d, \cdot)}_{n} = \text{softmax} (\boldsymbol{\delta}^{(d, \cdot)}_n)
    \label{eq:mixture-priors}
\end{equation}

\noindent where $\mathbf{M^{(d, \cdot)}}, \mathbf{C^{(d, \cdot)}} \in \mathbb{R}^{K}$ are learnable domain-specific parameters corresponding to the mean vector and diagonal covariance matrix.

After obtaining $\boldsymbol{\theta}^{(d, \cdot)}_{n}$, the expected rate of observing different "words" in cell \( n \) is parameterized by the following equation:

\begin{equation}
    \boldsymbol{r}^{(d,m)}_{n} = \text{softmax}\left(\boldsymbol{\theta}^{(d, \cdot)}_n \boldsymbol{\omega}^{(d, \cdot)} \boldsymbol{\rho}^{(\cdot, m)} + \boldsymbol{\lambda}^{(d,m)} \right)
\end{equation}

\noindent where $\boldsymbol{r}^{(d,m)}_{n}$ is $V^{(m)}$-dim vector, whose $v$-th element, $\boldsymbol{r}^{(d,m)}_{n, v}$,is the expected rate of observing feature \( v^{(m)} \) in cell \( n \). $\boldsymbol{\omega}^{(d, \cdot)} = \boldsymbol{\alpha} + \boldsymbol{\beta}^{(d, \cdot)}$ is composed of a shared term $\boldsymbol{\alpha}$ and a domain-specific term $\boldsymbol{\beta}^{(d, \cdot)}$. The shared term $\boldsymbol{\alpha}$ is a unified topic embedding to describe patterns across modalities and domains, while the additive term $\boldsymbol{\beta}^{(d, \cdot)}$ flexibly allows domain-specific variation in the topic features. $\boldsymbol{\rho}^{(\cdot, m)} \in \mathbb{R}^{L \times V^{(m)}}$ is the cross-domain feature embedding while $\boldsymbol{\lambda}^{(d,m)} \in \mathbb{R}^{1 \times V^{(m)}}$ is the scalar noise.

Finally, for each "read" of the $n$-th cell ("document") from the \(m\)-th modality and the \(d\)-th domain, a feature index is drawn from a categorical distribution represented by $r^{(d,m)}_n$. And $\mathbf{x}^{(d,m)}_{n, v}$ is the count of the word $v^{(m)}$. And express the data likelihood in terms of the read count :

\begin{equation}
    p(\boldsymbol{x}^{(d,m)}_{n}|\boldsymbol{r}^{(d,m)}_{n}) = \prod_{v=1}^{V^{(m)}} [\boldsymbol{r}^{(d,m)}_{n,v}]^{\boldsymbol{x}^{(d,m)}_{n,v}}
    \label{eq:likelihood}
\end{equation}

Based on the above generative process, in order to obtain modality-invariant representation $\mathbf{z}^{(d, \cdot)}_{n}$ for the cell $\mathbf{x}^{(d, \cdot)}_{n}$ , we use topic embedding $\boldsymbol{\omega}^{(d, \cdot)}$ and topic assignments $\boldsymbol{\theta}^{(d, \cdot)}_n$:
\begin{equation}
    \mathbf{z}^{(d, \cdot)}_{n} = \boldsymbol{\theta}^{(d, \cdot)}_n \boldsymbol{\omega}^{(d, \cdot)}
\end{equation}

\subsection{Model Design}
\label{sec:model-inference}

Based on the data generative process in Section \ref{sec:data-generative-process}, we design a variational model as illustrated in Figure \ref{fig:model}. We denote the parameter set of the model as $\Theta$. We aim to optimize $\Theta$ by maximizing the data log-likelihood over all cells across domains: $\mathcal{L} = \sum_{d=1}^D \sum_{n=1}^{N^{(d)}} \mathcal{L}^d_n$, where $\mathcal{L}^d_n$ is:  

\begin{equation}
    \begin{aligned}
        \mathcal{L}_n^d &= \log p^{(d, \cdot)}(\mathbf{x}^{(d, \cdot)}_{n} \mid \Theta) \\
        &= \log \int_{\boldsymbol{\delta}} p^{(d, \cdot)}(\mathbf{x}^{(d, \cdot)}_{n} \mid \boldsymbol{\delta}, \Theta) p^{(d, \cdot)}(\boldsymbol{\delta} | \Theta) d\mathbf{\boldsymbol{\delta}}
    \end{aligned}
    \label{eq:data-log-likelihood}
\end{equation}
\noindent where $p^{(d, \cdot)}\left(\boldsymbol{\delta} | \Theta\right)$ and  $p^{(d, \cdot)}(\mathbf{x}^{(d, \cdot)}_{n} \mid \boldsymbol{\delta}, \Theta)$ are parameterized by $\Theta$. In the following section, we will neglect $\Theta$ and write them as $p^{(d, \cdot)}(\boldsymbol{\delta})$ and $p^{(d, \cdot)}(\mathbf{x}^{(d, \cdot)}_{n} \mid \boldsymbol{\delta})$ for simplification. Since $\log p^{(d, \cdot)}(\mathbf{x}^{(d, \cdot)}_{n})$ is not tractable, we propose to maximize the evidence lower bound (ELBO) with the variational posterior $q^{(d, \cdot)} (\boldsymbol{\delta} \mid \mathbf{x}^{(d, \cdot)}_{n})$:



\begin{equation}
    \begin{aligned}
        & \log p^{(d, \cdot)}(\mathbf{x}^{(d, \cdot)}_{n})\\
        &= \log \int_{\boldsymbol{\delta}} p^{(d, \cdot)}(\mathbf{x}^{(d, \cdot)}_{n}\mid \boldsymbol{\delta}) p^{(d, \cdot)}(\boldsymbol{\delta})d\boldsymbol{\delta}\\
        &\geq \mathbb{E}_{q^{(d, \cdot)} (\boldsymbol{\delta} \mid \mathbf{x}^{(d, \cdot)}_{n})} \log p^{(d, \cdot)}(\mathbf{x}^{(d, \cdot)}_{n}\mid \boldsymbol{\delta}) \\
        &\quad \quad - \textsc{KL}(q^{(d, \cdot)} (\boldsymbol{\delta} \mid \mathbf{x}^{(d, \cdot)}_{n}) \parallel p^{(d, \cdot)}(\boldsymbol{\delta})) \\
        &= \mathbb{E}_{q^{(d, \cdot)} (\boldsymbol{\delta} \mid \mathbf{x}^{(d, \cdot)}_{n})} \sum_{m \in \mathbf{m}^{(d)}} \log p^{(d, \cdot)}(\mathbf{x}^{(d, m)}_{n}\mid \boldsymbol{\delta}) \\
        &\quad \quad - \textsc{KL}(q^{(d, \cdot)} (\boldsymbol{\delta} \mid \mathbf{x}^{(d, \cdot)}_{n}) \parallel p^{(d, \cdot)}(\boldsymbol{\delta})) \\
    \end{aligned}
    \label{eq:elbo}
\end{equation}
\noindent where we name such ELBO term for $\log p^{(d, \cdot)}(\mathbf{x}^{(d, \cdot)}_{n})$ as $ELBO^{(d, \cdot)}_{n}$. The full ELBO formulation is given in Appendix \ref{sec:appendix-loss}.

For $p^{(d, \cdot)}(\mathbf{x}^{(d, m)}_{n}\mid \boldsymbol{\delta})$ in this equation, we follow the data generative process in Section \ref{sec:data-generative-process} and parameterize $p^{(d, \cdot)}(\mathbf{x}^{(d, m)}_{n}\mid \boldsymbol{\delta})$ by $\{ \boldsymbol{\alpha}, \boldsymbol{\beta}^{(d, \cdot)}, \boldsymbol{\rho}^{(\cdot,m)}, \boldsymbol{\lambda}^{(d,m)} \}_{1 \leq d \leq D, m \in \mathbf{m}^{(d)}}$. This is corresponding to the "decoder" module in Figure \ref{fig:model}. 

For $q^{(d, \cdot)} (\boldsymbol{\delta} \mid \mathbf{x}^{(d, \cdot)}_{n})$, we estimate this posterior distributions as a normal distribution. To infer $q^{(d, \cdot)} (\boldsymbol{\delta} \mid \mathbf{x}^{(d, \cdot)}_{n})$ from multi-modality $\mathbf{x}^{(d, \cdot)}_{n}$, we model it as a product of normal distributions \cite{wu2018multimodal} across modality-specific topic assignment distributions within the domain. Particularly, we first model $q^{(d, \cdot)} (\boldsymbol{\delta} \mid \mathbf{x}^{(d, m)}_{n})$ as a normal distribution for each $m$ and compute its mean and covariance as follows:

\begin{equation}
    \label{eq:encoder}
    [ \boldsymbol{\mu}^{(d,m)}_{n}, \boldsymbol{\Sigma}^{(d,m)}_{n} ] = \text{NNET}^{(\cdot, m)} \left( \tilde{\mathbf{x}}^{(d,m)}_{n} ; \mathbf{W}^{(\cdot, m)} \right)
\end{equation}
\noindent where $\tilde{\mathbf{x}}^{(d,m)}_{n}$ is normalized counts for each feature as the raw count of the feature divided by the total counts of modality $m$ in cell $n$. $\text{NNET}^{(\cdot, m)}$ is a modality-specific neural network shared across domains with parameters denoted as $\mathbf{W}^{(\cdot, m)}$.




After obtaining $[ \boldsymbol{\mu}^{(d,m)}_{n}, \boldsymbol{\Sigma}^{(d,m)}_{n} ]$ for each available modality, $q^{(d, \cdot)} (\boldsymbol{\delta} \mid \mathbf{x}^{(d, \cdot)}_{n})$ is estimated as the product of these normal distributions. The mean and covariance of this joint normal for domain $d$ are computed as:
\begin{equation}
    {\boldsymbol{\mu}^{*}}^{(d, \cdot)}_n = \frac{\sum_{m \in \mathbf{m}^{(d)}} \boldsymbol{\mu}^{(d,m)}_{n} \boldsymbol{\Sigma}^{(d,m)}_{n}}{1 + \sum_{m \in \mathbf{m}^{(d)}} \boldsymbol{\Sigma}^{(d,m)}_{n}},    {\boldsymbol{\Sigma}^{*}}^{(d, \cdot)}_{n} = \frac{\prod_{m \in \mathbf{m}^{(d)}} \Sigma^{(d,m)}_{n}}{1 + \sum_{m \in \mathbf{m}^{(d)}} \Sigma^{(d,m)}_{n}}
    \label{eq:pog}
\end{equation}

The neural networks and the product-of-experts operation correspond to the "encoder" module in Figure \ref{fig:model}. Notably, our framework can flexibly handle variable modality availability in computing the topic assignment distribution, as the product is taken over available modalities $\mathbf{m}^{(d)}$ in domain $d$.



\subsection{Training Procedure}
\label{sec:train-pro}
With the model introduced in Section \ref{sec:model-inference}, the evidence lower bound (ELBO) in Equation \ref{eq:elbo} can be calculated as follows:

We resample $\boldsymbol{\delta}$ from $q^{(d, \cdot)} (\boldsymbol{\delta} \mid \mathbf{x}^{(d, \cdot)}_{n})$ and then compute the term $\mathbb{E}_{q^{(d, \cdot)} (\boldsymbol{\delta} \mid \mathbf{x}^{(d, \cdot)}_{n})} \sum_{m \in \mathbf{m}^{(d)}} \log p^{(d, \cdot)}(\mathbf{x}^{(d, m)}_{n}\mid \boldsymbol{\delta})$ by reparameterization trick \cite{kingma2013auto}, where $\log p^{(d, \cdot)}(\mathbf{x}^{(d, m)}_{n}\mid \boldsymbol{\delta})$ can be calculated via the procedure and equations in Section \ref{sec:model-inference}.

The KL-divergence in Equation (\ref{eq:elbo}) between the univariate Gaussian posterior $q^{(d, \cdot)} (\boldsymbol{\delta} \mid \mathbf{x}^{(d, \cdot)}_{n})$ and univariate Gaussian learnable priors $p^{(d, \cdot)} (\boldsymbol{\delta})$ for a domain $d$, which we express as $\textsc{KL}\left(q (\boldsymbol{\delta}) \parallel p_\theta(\boldsymbol{\delta})\right)$ for simplicity, has a closed-form solution: 

\begin{equation}
    \begin{aligned}
        \textsc{KL}\left(q (\boldsymbol{\delta}) \parallel p(\boldsymbol{\delta})\right) = \log\left(\frac{\sqrt{\boldsymbol{\Sigma}_p}}{\sqrt{\boldsymbol{\Sigma}_q}}\right) + \frac{\boldsymbol{\Sigma}_q + (\boldsymbol{\mu}_q - \boldsymbol{\mu}_p)^2}{2\boldsymbol{\Sigma}_p} - \frac{1}{2}
    \end{aligned}
    \label{eq:kl-caluclation}
\end{equation}
\noindent where the full derivation is given in Appendix \ref{sec:appendix-loss}.

All the parameters $\Theta$ of the model, including the decoder weights $\{\boldsymbol{\alpha}, \boldsymbol{\beta}^{(d, \cdot)}, \boldsymbol{\rho}^{(\cdot,m)}, \boldsymbol{\lambda}^{(d,m)} \}_{1 \leq d \leq D, m \in \mathbf{m}^{(d)}}$
and the encoder weights $\{\mathbf{W}^{(\cdot, m)}\}_{1 \leq m \leq M}$ are jointly optimized by minimizing loss $\mathcal{L}$ across $D$ domains:
\begin{equation}
    \Theta \leftarrow \argmin_{\Theta} \sum_{d=1}^D \mathcal{L}^{(d, \cdot)}
    \label{eq:train-updates}
\end{equation}
\noindent where $\mathcal{L}^{(d, \cdot)}$ is computed as follows:
\begin{equation}
    \mathcal{L}^{(d, \cdot)} = - \frac{1}{N^{(d, \cdot)}} \sum_{n=1}^{N^{(d, \cdot)}} ELBO^{(d, \cdot)}_{n} + \lambda_{\beta} \|\boldsymbol{\beta}^{(d, \cdot)}\|_2
    \label{eq:final-objective-wo-ncl}
\end{equation}

Here we also constrain the norm of $\boldsymbol{\beta}$ to ensure that $\alpha$ is representative of global topics and $\lambda_{\beta}$ is the weight.

\subsection{Neighborhood Contrastive Loss }
\label{sec:model-ncl}


One issue that may arise in the missing modality setting is the bias to certain modalities. For example, the embedding may rely on features from modalities that are available across the majority of the domains, leading to poor performance in some domains where this modality is unavailable. Therefore, we ensure that available modalities in each domain $d$ are equally represented by regularizing the learned embedding $\boldsymbol{\delta}^{(d, \cdot)}$ to preserve cell-cell relations for all cells in domain $d$.


For two different cells $\mathbf{x}^{(d, m)}_{n}, \mathbf{x}^{(d, m)}_{i}$ from the same domain $d$  and modality $m \in \mathbf{m}^{(d)}$, we measure their distance as the Euclidean distance between $\mathbf{x}^{(d, m)}_{n}$ and $\mathbf{x}^{(d, m)}_{n}$. Then the k-nearest neighbor
samples for a given sample $\mathbf{x}^{(d, \cdot)}_{n}$ on modality $m$ can be computed with this distance metric, which is denoted as $\mathbf{nn}^{(d, m)}_k(n)$. Note that $\mathbf{nn}^{(d, m)}_k(n)$ might be different across modalities.

The neighborhood contrastive loss (NCL) objective maximizes the similarity between cell topic features from cells and their nearest neighboring cells in domain $d$ and modality $m$, while minimizing the embedding similarity of cells from separate neighborhoods:

\begin{equation}
    \mathcal{L}^{(d, \cdot)}_{cont}(n) = - \textstyle \sum_{m \in \mathbf{m}^{(d)}}\sum_{i \in \mathbf{nn}^{(d, m)}_k(n)} \log \left( \frac{\sigma(\boldsymbol{\delta}_n^{(d, m)}, \boldsymbol{\delta}_i^{(d, m)})}{ \sum_{j \neq n} \sigma(\boldsymbol{\delta}_n^{(d, m)}, \boldsymbol{\delta}_j^{(d, m)})} \right)
    \label{eq:contrastive-loss}
\end{equation}

\noindent where embedding similarity $\sigma(\boldsymbol{\delta}_n^{(d, \cdot)}, \boldsymbol{\delta}_i^{(d, \cdot)})$ between cells $n$ and $i$ is computed as:
\begin{equation}
    \sigma(\boldsymbol{\delta}_n^{(d, m)}, \boldsymbol{\delta}_i^{(d, m)}) = \text{exp}\left( \frac{\langle\boldsymbol{\delta}_n^{(d, m)}, \boldsymbol{\delta}_i^{(d, m)}\rangle}{\kappa \|\boldsymbol{\delta}_n^{(d, m)}\| \cdot \|  \boldsymbol{\delta}_i^{(d, m)}\|} \right)
\end{equation}
\noindent for a positive constant $\kappa$.

Once we incorporate this contrastive loss, the final objective function becomes:
\begin{equation}
    \mathcal{L}^{(d, \cdot)} = - \frac{1}{N^{(d, \cdot)}} \sum_{n=1}^{N^{(d, \cdot)}} ELBO^{(d, \cdot)}_{n} + \lambda_{\beta} \|\boldsymbol{\beta}^{(d, \cdot)}\|_2 + \frac{1}{N^{(d)}} \sum_{n=1}^{N^{(d)}} \mathcal{L}^{(d, \cdot)}_{cont}(n)
    \label{eq:final-objective}
\end{equation}

\subsection{Model Inference}
\label{sec:test-time-inference}
For any sample $\mathbf{x}^{(d, \cdot)}_{*}$ in domain $d$, we can compute its integrated feature representation and impute missing modalities.

\subsubsection*{Integration Representation} We first feed the features $\mathbf{x}^{(d, \cdot)}_{*}$ to the encoders and aggregate the result as in Eq. (\ref{eq:encoder})-(\ref{eq:pog}) to obtain the cell topic embedding $\boldsymbol{\delta}_n^{(d, \cdot)}$. We then compute the topic mixture probability $\boldsymbol{\theta}_n^{(d, \cdot)}$ by applying a softmax transformation on $\boldsymbol{\delta}_n^{(d, \cdot)}$ to obtain the final integrated feature representation:

\begin{equation}
    \mathbf{z}^{(d, \cdot)}_{*}= \boldsymbol{\theta}_n^{(d, \cdot)} (\boldsymbol{\alpha} + \boldsymbol{\beta}^{(d,\cdot)})
\end{equation}

\subsubsection*{Missing Modality Imputation} Our framework supports missing modality imputation, \textit{even if a modality is missing across all samples in a domain}. We can impute the missing features $\hat{\mathbf{X}}^{(i,\bar{m})}$ for any $\bar{m} \notin \mathbf{m}^{(d)}$ as follows:
\begin{equation}
    \hat{\mathbf{x}}^{(d,\bar{m})}_{n}= \text{softmax}\left(\boldsymbol{\theta}_n^{(d, \cdot)} (\boldsymbol{\alpha} + \boldsymbol{\beta}^{(d,\cdot)})\boldsymbol{\rho}^{(\cdot, \bar{m})}\right)
    \label{eq:imputation}
\end{equation}

\section{Experiments}

\subsection{Dataset and Experiment Settings}
We use the 2021 NeurIPS single-cell challenge dataset with inherent missing modality settings (unmeasured modalities) and simulated missing modality settings (masked modalities) to evaluate the methods \cite{luecken2021sandbox}. This dataset contains bone marrow mononuclear cells profiled Multiome (gene expression \& chromatin accessibility) \cite{buenrostro2015single} and CITE-seq (gene expression \& protein abundance) \cite{stoeckius2017simultaneous}, respectively.

The NeurIPS dataset is suitable for evaluating integration robustness under missing modality settings as it inherently contains missing modalities (chromatin accessibility is missing from CITE-seq samples while protein abundance is missing from Multiome samples). In order to evaluate the imputation performance, we must simulate missing modalities where we have access to ground-truth features.

We consider three different scenarios with the NeurIPS single-cell challenge dataset, as shown in Figure \ref{fig:exp_sce}. We select half of the domains in each experimental scenario as \textit{imputation target domains}, where we simulate all imputation possibilities using our ground-truth features. We formulate these scenarios as follows:

\begin{figure}
    \begin{center}
        \includegraphics[scale=0.45]{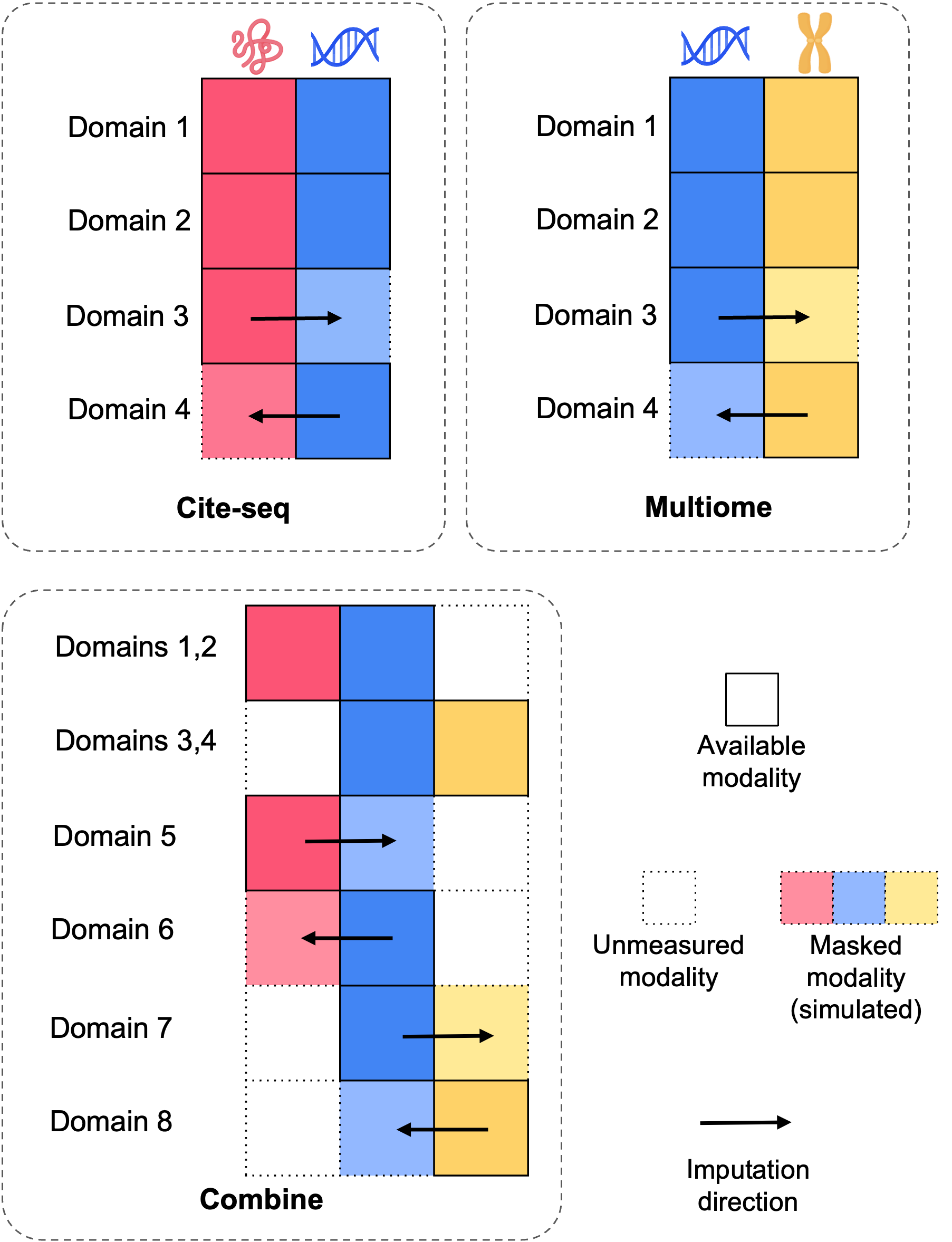}
    \end{center}
    \caption{Domains and modality availability under each experiment scenario.}
    \label{fig:exp_sce}
    \vspace{-1.5em}
\end{figure}

\begin{enumerate}

    \item \textit{CITE-seq.} We consider the data of CITE-seq from four different sites. We treat each site as one domain. In this setting, two modalities (gene expression (GEX) \& protein abundance (ADT)) are considered. We assume the availability of both modalities in domains 1 and 2, protein abundance in domain 3 and gene expression in domain 4.
    \item \textit{Multiome.} We consider the data of Multiome from four different sites. We treat each site as one domain. In this setting, two modalities (gene expression (GEX) \& chromatin accessibility (ATAC)) are considered. We assume the availability of both modalities in domains 1 and 2, gene expression in domain 3 and protein abundance in domain 4.
    \item \textit{Combine.} We combine the above data of CITE-seq and Multiome. We treat each site as one domain. In this setting, three modalities (GEX, ADT, ATAC) are considered. We assume the availability of gene expression \& protein abundance in domains 1 and 2, gene expression \& chromatin accessibility in domains 3 and 4. For domains 5-8, only one modality is available.
\end{enumerate}


\begin{table}[thbp]
	\caption{Results of embedding-based clustering. Two metrics ARI and NMI are used to evaluate the performance. Averaged performance over different imputation target domains are reported.}
	\small
    \renewcommand{\arraystretch}{1.5}
	\begin{tabular}{ccccc}
        \toprule
        &  Method & Cite-seq & Multiome & Combine \\
        \hline
        \multirow{3}{*}{ARI} & CCA & $0.497 $ & $0.432$ & $0.360$\\
        & moETM  & $0.546$ & $0.561$ & $0.522$\\
        & \methodnamerm~  & $\mathbf{0.590}$ & $\mathbf{0.578}$ & $\mathbf{0.572}$\\
        \hline
        \multirow{3}{*}{NMI} & CCA  & $0.694$ & $0.307$ & $0.523$\\
        & moETM  & $0.731$ & $0.632$ & $0.693$\\
        & \methodnamerm~  & $\mathbf{0.746}$ &  $\mathbf{0.641}$ & $\mathbf{0.706}$\\
        \bottomrule
    \end{tabular}
    \label{table:cluster}
    \vspace{-1.0em}
\end{table}

\begin{table}[thbp]
	\caption{Results of cell type classification. Accuracy is used to evaluate the performance. Averaged performance over different imputation target domains are reported.}
	\small
    \renewcommand{\arraystretch}{1.5}
	\begin{tabular}{cccc}
        \toprule
        Method & Cite-seq & Multiome & Combine \\
        \hline
        CCA & $0.657$ & $0.589$ & $0.539$\\
        moETM  & $\mathbf{0.796}$ & $0.637$ & $0.748$\\
        \methodnamerm~ & $0.794$ & $\mathbf{0.671}$ & $\mathbf{0.751}$\\
        \bottomrule
    \end{tabular}
    \label{table:classification}
    \vspace{-1.0em}
\end{table}

\begin{figure*}
    \begin{center}
    \includegraphics[scale=0.33]{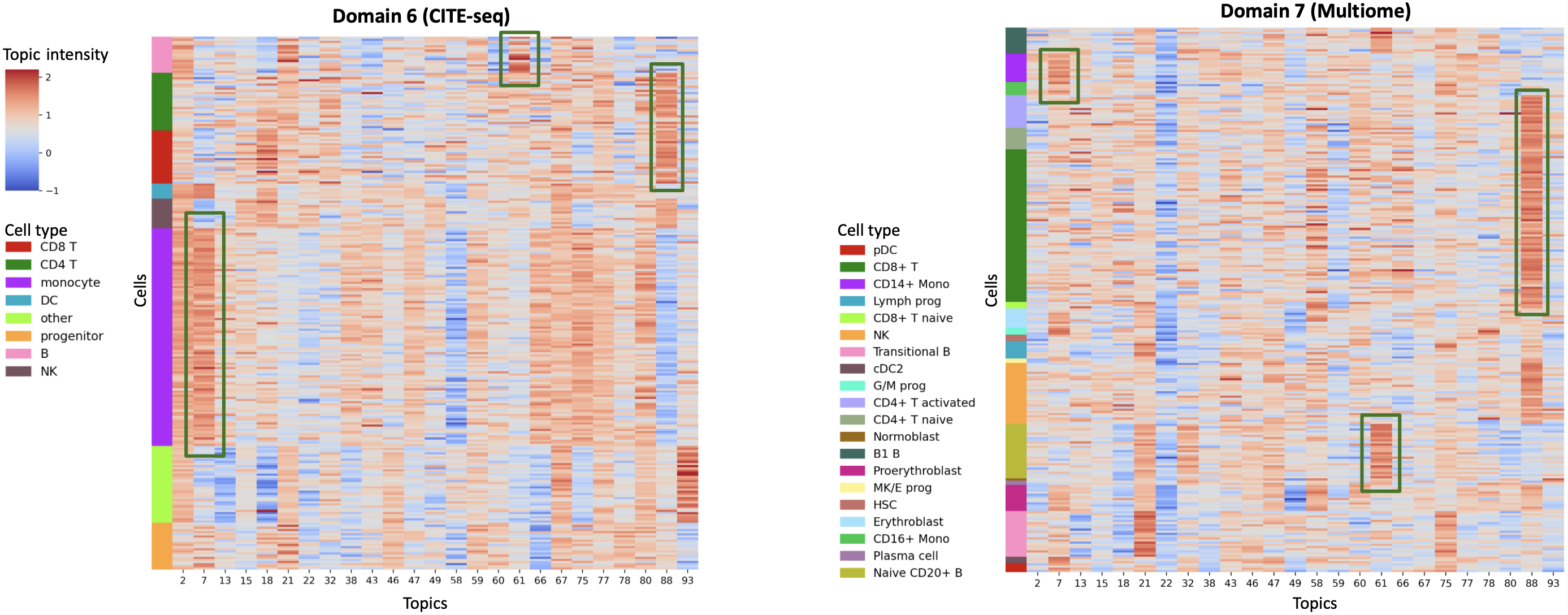}
    \end{center}
    \caption{Distribution of topic assignment scores in the $Combine$ setting. The top 20\% of assigned topics are selected across 500 sampled cells from 2 domains. Color intensity values correspond to the cell-topic feature value before normalization into a topic mixture probability. Boxed topic features correspond to cell topic features with strong association to unique cell types.}
    \label{fig:cell-topics-heatmap}
    \vspace{-1.5em}
\end{figure*}

\subsection{Evaluation Metric}

We evaluate the performance of all methods from two aspects:

\subsubsection{Integrated feature quality from dataset combination}
\label{subsec:integration-results}
 We follow previous works \cite{zhou2023moetm} and assess the self-supervised multi-modal integration using the following downstream tasks: 
 
\begin{enumerate}
    \item \textit{Self-supervised cell type clustering.} We apply k-means clustering on the integrated multi-modal embeddings to cluster cells into biologically meaningful clusters that are useful for identifying cell types. We evaluate clustering quality using adjusted Rand index (ARI) and normalized mutual information (NMI) between k-means cell clusters and cell type labels.
    \item \textit{Supervised cell type classification.} We also use the integrated embeddings to train a downstream cell type classifier. We train the classifier on the integrated cell embeddings and evaluate the performance on a hold-out set using cross-entropy.
\end{enumerate}

\subsubsection{Imputation quality under missing modality settings}
\label{subsec:imputation-results}
We assess the imputation quality under simulated missing modality settings using  Pearson correlations between the true and imputed features.

\subsection{Baselines}
We compare \methodname~ with the following baseline methods by extending them to our proposed setting where \textit{no full-modality reference is available.} As previous single-cell integration methods are unable to integrate features across cohorts/domains with conflicting modality availability in this proposed setting, we use modified baseline methods that are suited for our task.

\begin{enumerate}
    \item \textbf{"Missing-aware" CCA} is an extension of canonical correlation analysis (CCA), which has been used for single-cell integration \cite{butler2018integrating}, that produces multi-modal embeddings using SVD across samples while masking missing samples.
    \item \textbf{"Missing-aware" moETM} is an extension of a multi-omic topic embedding framework (moETM) \cite{zhou2023moetm} that learns multi-modal embeddings using a variational autoencoder across samples while masking missing samples.
\end{enumerate}

Both of these modified baselines can be used to learn integrated representation after adjusting for missing modalities, while "missing-aware" moETM supports imputing missing modalities. Therefore, we compare \methodname~ with both baselines on cell type clustering and classification and compare \methodname~ with moETM on missing modality imputation.

\subsection{Cell Type Clustering and Classification}
\label{sec:clustering-and-classification}

\begin{figure*}
    \begin{center}
        \includegraphics[scale=0.038]{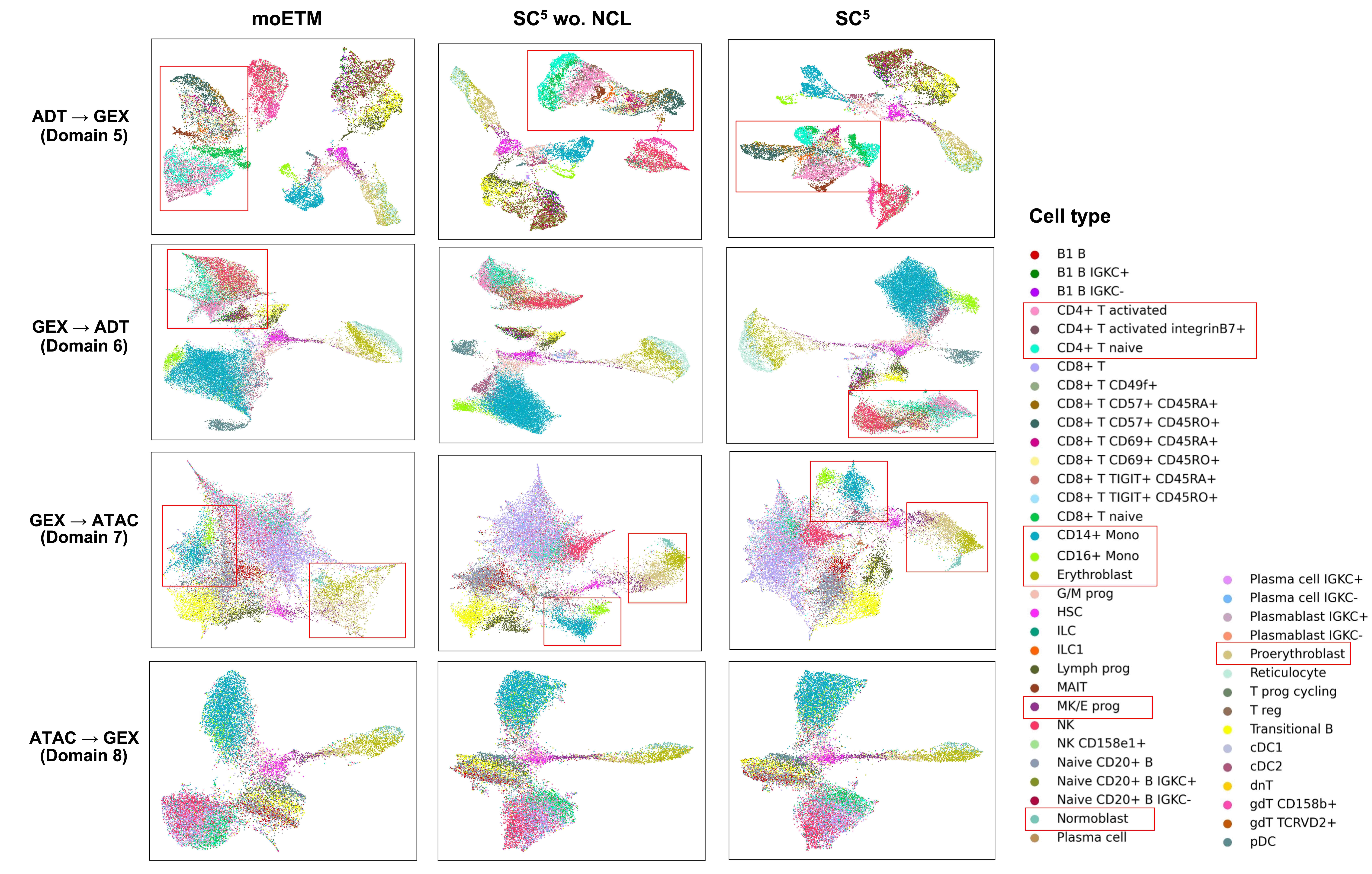}
    \end{center}
    \caption{UMAP visualization of cell clustering on imputed modalities under the $Combine$ setting. Each row corresponds the imputation results in one domain. In each subfigure, individual data points represent cells, while different colors are used to denote different cell types.}
    \label{fig:umap}
    \vspace{-1.0em}
\end{figure*}

The experimental results on cell type clustering and classification are shown in Tables \ref{table:cluster} and \ref{table:classification}. Topic-modelling based methods outperforms CCA by large margin, which indicates that topic modelling is more suitable for describing cell activities. \methodname~ outperforms the baselines in all but one of the settings. As NMI is best suited for evaluating imbalanced clusters, which we observe in practice (Figure \ref{fig:umap}), this metric is higher than ARI across all methods.

We further demonstrate that \methodname~ learns informative topic features $\boldsymbol{\omega}$ for identifying cell types by visualizing the topic distribution. In Figure \ref{fig:cell-topics-heatmap}, we visualize the distribution of the top 20\% of inferred topics shared among 500 sampled cells across 2 domains. We annotate cell topic features that have high topic scores for unique cell types. Notably, \methodname~ infers common topics from separate domains that are associated with the same cell type. For example, topic 7 is associated with high topic intensity for monocyte cells in both domain 6 (annotated as `monocyte') and in domain 7 (annotated by the subtypes `CD16+ Mono' and `CD14+ Mono'). Additionally, topic 61 is associated with `B' cells in both domains.

In Appendix \ref{appendix:topic-embedding-analysis}, we compare the learned topic embeddings between domains and find that the integrated features are flexible to domain-dependent heterogeneity. These results suggest that \methodname~ learns topic features that are informative while being flexible to domain-specific heterogeneity.

\begin{table}[thbp]
	\caption{Results of missing modality imputation. Pearson correlation  between true and imputed features is used to evaluate the performance. Averaged performance over different imputation target domains are reported.}
	\small
    \renewcommand{\arraystretch}{1.5}
	\begin{tabular}{cccc}
        \toprule
        Method & Cite-seq & Multiome & Combine\\
        \hline
        moETM  & $0.465$ & $0.184$ & $0.326$\\
        \methodnamerm~ & $\mathbf{0.671}$ & $\mathbf{0.278}$ & $\mathbf{0.444}$ \\
        \bottomrule
    \end{tabular}
    \label{table:imp}
    \vspace{-1.0em}
\end{table}

\subsection{Missing Modality Imputation}

The experimental results on missing modality imputation are displayed in Table \ref{table:imp}. From Table \ref{table:imp}, \methodname~ shows a significant performance improvement over the baselines.

By the results, we demonstrate that \methodname~ can impute  modalities that are completely missing in  a domain, even when a full multimodal reference is unavailable during training. By learning complex relationships between samples from varying modalities and domains, our method can impute features that match the modality-specific and domain-specific distributions of the available data. Unlike previous works, \methodname~ can impute features for a modality that is completely unseen in a target dataset, as long as the target modality is available in at least one domain during training.

\begin{table}[thbp]
	\caption{Ablation study on neighborhood contrastive loss. Averaged performance over different imputation target domains are reported.}
	\small
    \renewcommand{\arraystretch}{1.5}
	\begin{tabular}{ccccc}
        \toprule
        &  Method & Cite-seq & Multiome & Combine \\
        \hline
        \multirow{2}{*}{ARI} & \methodnamerm~ wo. NCL & $0.529$ & $0.523$ & $0.554$\\
        & \methodnamerm~  & $\mathbf{0.590}$ & $\mathbf{0.578}$ & $\mathbf{0.572}$\\
        \hline
        \multirow{2}{*}{NMI} & \methodnamerm~ wo. NCL & $0.738$ & $\mathbf{0.643}$ & $0.699$\\
        & \methodnamerm~  & $\mathbf{0.746}$ &  $0.641$ & $\mathbf{0.706}$\\
        \hline
        \multirow{2}{*}{Classification} & \methodnamerm~ wo. NCL & $\mathbf{0.804}$ & $\mathbf{0.699}$ & $\mathbf{0.757}$ \\
        & \methodnamerm~  & $0.794$ & $0.671$ & $0.751$\\
        \hline
        \multirow{2}{*}{Imputation} & \methodnamerm~ wo. NCL & $\mathbf{0.694}$ & $\mathbf{0.286}$ & $0.435$ \\
        & \methodnamerm~ & $0.671$ & $0.278$ & $\mathbf{0.444}$ \\
        \bottomrule
    \end{tabular}
    \label{table:aba}
    \vspace{-1.0em}
\end{table}

We verify \methodname's imputation capabilities by visualizing the imputed features using Uniform Manifold Approximation and Projection (UMAP) \cite{mcinnes2018umap} in Figure \ref{fig:umap}. Our \methodname~ variants impute features that preserve finer-grained cell type clusters in contrast to moETM which does not explicitly  model domain-dependent features. In particular, "CD4+ T cells" associated with immune response are have greater disentaglement when imputing protein abundance from gene expression (rows 1, 2). When imputing chromatin accessibility from gene expression (row 3), \methodname~ better separates "Erythoroblast", "Proerythroblast", "Normoblast", and "MK/E prog" cells associated with red blood cell production. We note that on the most difficult imputation task (imputing dense gene expression features from sparse chromatin accessibility features), the imputed cell features share similar patterns, though \methodname~ produces embeddings that are more compact compared to the baseline. Our results  demonstrate that \methodname~ learns robust domain-dependent and modality factors that preserve cell type clustering patterns when imputing masked features.

\subsection{Ablation Study}
We now experimentally validate the effectiveness of neighborhood contrastive loss (NCL) introduced in Section \ref{sec:model-ncl}. We conduct an ablation study by comparing \methodname~ with its variant obtained by removing the neighborhood contrastive loss term. As the results show in Table \ref{table:aba}, adding the neighborhood contrastive loss (NCL) term consistently improves the clustering performance. Notably, the version with neighborhood contrastive loss can achieve the best or comparable classification and imputation results for all tasks under the $Combine$ setting.

\section{Conclusion}

In this paper, we propose a novel framework for joint analysis of single-cell data across cohorts (domains) with incomplete modalities. The proposed framework utilizes topic modeling to capture the data generative process, incorporating shared and domain-specific parameters. Through experiments on real-world datasets, we demonstrate that our proposed framework can learn integrated representations and impute missing modalities more efficiently than the baselines.

\bibliographystyle{ACM-Reference-Format}
\bibliography{ref}

\appendix
\section{Data processing}
\label{sec:appendix-data-processing}
All datasets were processed into the format of cells-by-features matrices. 
For each dataset, read counts for each feature were first normalized per cell by total counts within the same omic. This normalization was executed using the \textit{scanpy}. Subsequently, a log1p transformation was applied to the normalized data.

Following normalization and transformation, we identified highly variable features by \textit{scanpy.pp.highly\_variable\_genes}. In the case of protein features, all measured surface proteins were included due to the inherently limited quantity of proteins measured by the scADT-seq assay in comparison to genes or chromatin regions, and their substantial elucidation of cellular functions.

\section{Derivations for Loss Functions}
\label{sec:appendix-loss}
\subsection{Evidence Lower Bound}
\begin{equation}
    \begin{aligned}
        & \log p^{(d, \cdot)}(\mathbf{x}^{(d, \cdot)}_{n})\\
        &= \log \int_{\boldsymbol{\delta}} p^{(d, \cdot)}(\mathbf{x}^{(d, \cdot)}_{n}\mid \boldsymbol{\delta}) p^{(d, \cdot)}(\boldsymbol{\delta})d\boldsymbol{\delta}\\
        &= \log \int_{\boldsymbol{\delta}} q^{(d,\cdot)} (\boldsymbol{\delta} \mid \mathbf{x}^{(d, \cdot)}_{n})) \frac{ p^{(d, \cdot)}(\mathbf{x}^{(d, \cdot)}_{n}\mid \boldsymbol{\delta}) p^{(d, \cdot)}(\boldsymbol{\delta})}{q^{(d, \cdot)} (\boldsymbol{\delta} \mid \mathbf{x}^{(d, \cdot)}_{n})} d\boldsymbol{\delta} \\
        &=  \log \mathbb{E}_{q^{(d, \cdot)} (\boldsymbol{\delta} \mid \mathbf{x}^{(d, \cdot)}_{n})} \frac{ p^{(d, \cdot)}(\mathbf{x}^{(d, \cdot)}_{n}\mid \boldsymbol{\delta}) p^{(d, \cdot)}(\boldsymbol{\delta})}{q^{(d, \cdot)} (\boldsymbol{\delta} \mid \mathbf{x}^{(d, \cdot)}_{n})} \\
        &\geq  \mathbb{E}_{q^{(d, \cdot)} (\boldsymbol{\delta} \mid \mathbf{x}^{(d, \cdot)}_{n})} \log \frac{ p^{(d, \cdot)}(\mathbf{x}^{(d, \cdot)}_{n}\mid \boldsymbol{\delta}) p^{(d, \cdot)}(\boldsymbol{\delta})}{q^{(d, \cdot)} (\boldsymbol{\delta} \mid \mathbf{x}^{(d, \cdot)}_{n})} \\
        &= \mathbb{E}_{q^{(d, \cdot)} (\boldsymbol{\delta} \mid \mathbf{x}^{(d, \cdot)}_{n})} \log p^{(d, \cdot)}(\mathbf{x}^{(d, \cdot)}_{n}\mid \boldsymbol{\delta}) \\
        &\quad \quad - \textsc{KL}(q^{(d, \cdot)} (\boldsymbol{\delta} \mid \mathbf{x}^{(d, \cdot)}_{n}) \parallel p^{(d, \cdot)}(\boldsymbol{\delta})) \\
        &= \mathbb{E}_{q^{(d, \cdot)} (\boldsymbol{\delta} \mid \mathbf{x}^{(d, \cdot)}_{n})} \sum_{m \in \mathbf{m}^{(d)}} \log p^{(d, \cdot)}(\mathbf{x}^{(d, m)}_{n}\mid \boldsymbol{\delta}) \\
        &\quad \quad - \textsc{KL}(q^{(d, \cdot)} (\boldsymbol{\delta} \mid \mathbf{x}^{(d, \cdot)}_{n}) \parallel p^{(d, \cdot)}(\boldsymbol{\delta})) \\
    \end{aligned}
\end{equation}

\subsection{KL-Divergence}
The KL-divergence in Equation \ref{eq:elbo} between the univariate Gaussian posterior $q^{(d, \cdot)}(\boldsymbol{\delta} \mid \mathbf{x})$ and univariate Gaussian learnable priors $p_\theta^{(d, \cdot)} (\boldsymbol{\delta})$ for a domain $d$, which we express as $\textsc{KL}\left(q (\boldsymbol{\delta}) \parallel p_\theta(\boldsymbol{\delta})\right)$ for simplicity, is derived as follows:
\begin{equation}
    \begin{aligned}
        \textsc{KL}\left(q (\boldsymbol{\delta}) \parallel p_\theta(\boldsymbol{\delta})\right) &= \int q (\boldsymbol{\delta}) \log \frac{q(\boldsymbol{\delta})}{p_\theta(\boldsymbol{\delta})} d\boldsymbol{\delta} \\
        &= \int q(\boldsymbol{\delta}) \left[ \log q(\boldsymbol{\delta}) - \log p_\theta(\boldsymbol{\delta}) \right] d\boldsymbol{\delta} \\
        &= \int \frac{1}{\sqrt{2\pi \boldsymbol{\Sigma}_q}}\text{exp}\left[\frac{(\boldsymbol{\delta}-\boldsymbol{\mu}_q)^2}{2\boldsymbol{\Sigma}_q} \right] \\
        & \quad \times \left[\log \frac{\sqrt{\boldsymbol{\Sigma}_p}}{\sqrt{\boldsymbol{\Sigma}_q}} - \left( \frac{(\boldsymbol{\delta}-\boldsymbol{\mu}_p)^2}{2\boldsymbol{\Sigma}_p} +\frac{(\boldsymbol{\delta}-\boldsymbol{\mu}_q)^2}{2\boldsymbol{\Sigma}_q}  \right) \right] \\
        &= \mathbb{E}_{q(\boldsymbol{\delta})} 
        \left[ \log \frac{\sqrt{\boldsymbol{\Sigma}_p}}{\sqrt{\boldsymbol{\Sigma}_q}} - \left( \frac{(\boldsymbol{\delta}-\boldsymbol{\mu}_p)^2}{2\boldsymbol{\Sigma}_p} + \frac{(\boldsymbol{\delta}-\boldsymbol{\mu}_q)^2}{2\boldsymbol{\Sigma}_q}  \right) \right] \\
        &= \log \frac{\sqrt{\boldsymbol{\Sigma}_p}}{\sqrt{\boldsymbol{\Sigma}_q}}   + \mathbb{E}_{q(\boldsymbol{\delta})} \left[ \frac{(\boldsymbol{\delta}-\boldsymbol{\mu}_p)^2}{2\boldsymbol{\Sigma}_p} \right] -\frac{1}{2} \\
        &= \log \frac{\sqrt{\boldsymbol{\Sigma}_p}}{\sqrt{\boldsymbol{\Sigma}_q}} +  \frac{\boldsymbol{\Sigma}_p + (\boldsymbol{\mu}_q-\boldsymbol{\mu}_p)^2}{2 \boldsymbol{\Sigma}_p}-\frac{1}{2}
    \end{aligned}
    \label{eq:kl-caluclation}
\end{equation}

\section{Topic embedding analysis}
\begin{figure}[!hb]
    \includegraphics[scale=0.11]{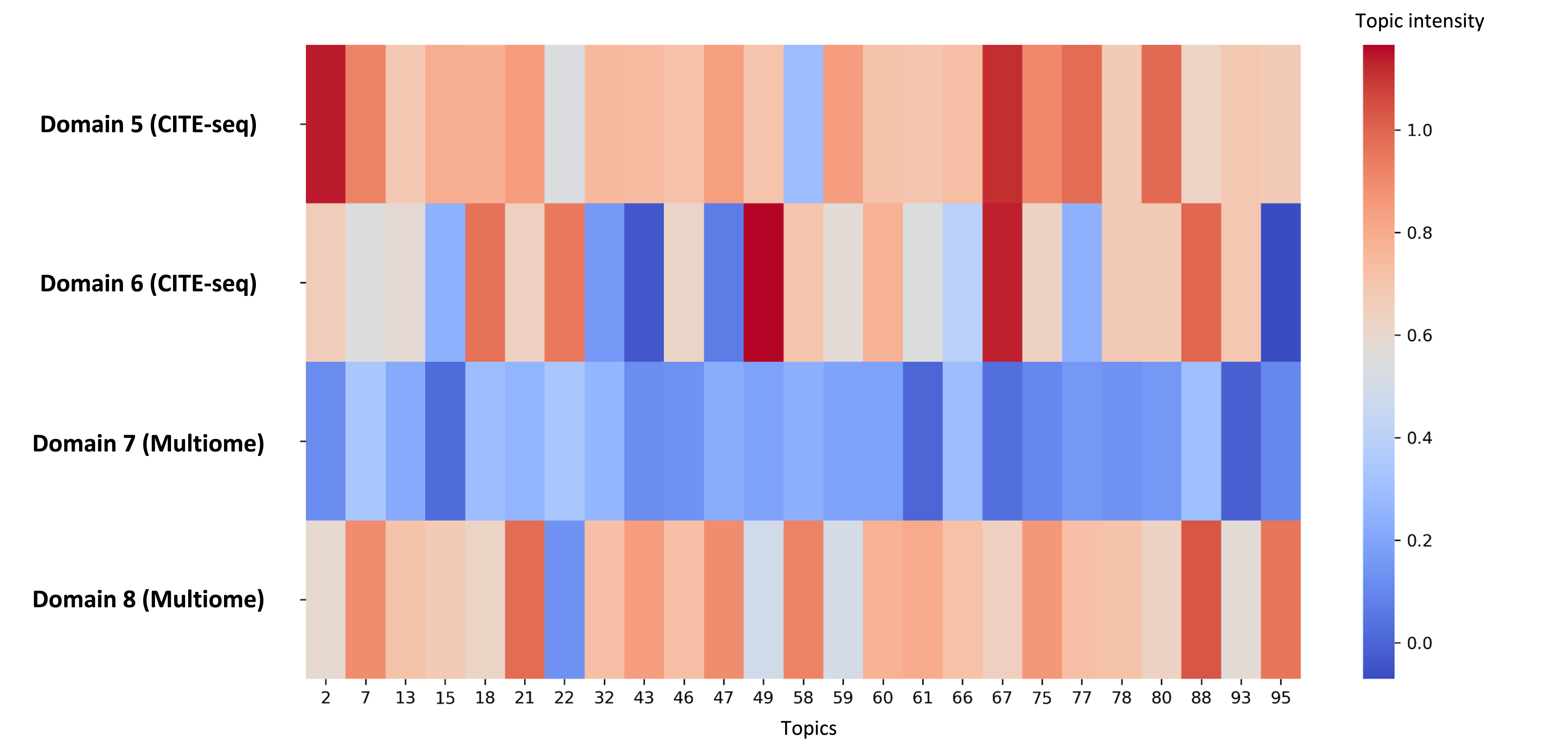}
    \caption{Distribution of scores for top 20\% of assigned topics averaged across samples within select domains. Topic features are learned by under the \textit{Combine} setting.}
    \label{fig:domain-topics-heatmap}
\end{figure}

We assess \methodname's ability to infer latent topic features that are flexible to domain-specific heterogeneity. In figure \ref{fig:domain-topics-heatmap}, we visualize the distribution of the top 20\% of topics inferred by \methodnamerm~ averaged within each test domain. Notably, our method is able to learn cell topic feature distributions that are unique to each domain.

\label{appendix:topic-embedding-analysis}

\end{document}